\pdfoutput=1

\documentclass[11pt]{article}

\usepackage{emnlp2021}

\usepackage{times}
\usepackage{latexsym}
 \usepackage{balance}

\usepackage[T1]{fontenc}

\usepackage[utf8]{inputenc}

\usepackage{microtype}
\usepackage{amsmath}
\usepackage{graphicx}
\usepackage{subcaption}
\usepackage{booktabs}
\usepackage{adjustbox}
\usepackage{ulem}
\usepackage{multirow}
\usepackage[justification=justified, skip=5pt]{caption}

\usepackage{soul}
\sethlcolor{tea_green}
\usepackage{color}
\definecolor{tea_green}{RGB}{214, 234, 193}
\definecolor{hint_green}{RGB}{226,246,209}
\definecolor{Madang}{RGB}{190,235,159}
\definecolor{yellow_green}{RGB}{198,222,119}
\definecolor{link_water}{RGB}{221, 232, 250}
\definecolor{celestial_blue}{RGB}{52, 152, 219}
\definecolor{shakespeare}{RGB}{85, 154, 193}
\definecolor{buttermilk}{RGB}{255,242,174}
\definecolor{chardonnay}{RGB}{250,196,114}
\definecolor{rajah}{RGB}{253,180,98}
\definecolor{fog}{RGB}{213, 193, 234}
\definecolor{melon}{RGB}{254,191,181}
\definecolor{sundown}{RGB}{249, 180, 181}
\definecolor{mona_lisa}{RGB}{246,152,134}
\definecolor{salmon}{RGB}{242,131,107}

\definecolor{saltpan}{RGB}{238, 243, 232}
\definecolor{aqua_spring}{RGB}{232, 243, 232}
\definecolor{tea_green}{RGB}{214, 234, 193}
\definecolor{Madang}{RGB}{190,235,159}
\definecolor{fringy_flower}{RGB}{194, 234, 193}
\definecolor{aero_blue}{RGB}{193, 234, 213}
\definecolor{pixie_green}{RGB}{183,214,170}
\definecolor{french_pass}{RGB}{195,232,246}
\definecolor{ice_cold}{RGB}{169,232,220}
\definecolor{pale_turquoise}{RGB}{172,240,242}
\definecolor{cruise}{RGB}{179,226,205}
\definecolor{sail}{RGB}{163,205,235}
\definecolor{spindle}{RGB}{179,205,227}
\definecolor{link_water}{RGB}{221, 232, 250}
\definecolor{periwinkle}{RGB}{203,213,232}
\definecolor{zanah}{RGB}{220, 233, 213}
\definecolor{frostee}{RGB}{217, 231, 214}
\definecolor{opal}{RGB}{199, 221, 211}
\definecolor{jet_stream}{RGB}{188, 214, 210}
\definecolor{skeptic}{RGB}{153, 187, 167}
\definecolor{hint_green}{RGB}{226,246,209}
\definecolor{snow_flurry}{RGB}{230,245,201}
\definecolor{surf_crest}{RGB}{205,230,208}
\definecolor{yellow_green}{RGB}{198,222,119}
\definecolor{cream}{RGB}{255,255,204}
\definecolor{pale_prim}{RGB}{255,255,179}
\definecolor{spring_sun}{RGB}{242,243,195}
\definecolor{portafino}{RGB}{245,237,160}
\definecolor{buttermilk}{RGB}{255,242,174}
\definecolor{cream_brulee}{RGB}{255, 229, 151}
\definecolor{dairy_cream}{RGB}{254,226,189}
\definecolor{champagne}{RGB}{254,217,166}
\definecolor{chardonnay}{RGB}{250,196,114}
\definecolor{manhattan}{RGB}{226,180,125}
\definecolor{rajah}{RGB}{253,180,98}
\definecolor{early_dawn}{RGB}{252,243,218}
\definecolor{egg_shell}{RGB}{238, 234, 215}
\definecolor{selago}{RGB}{243, 232, 243}
\definecolor{quartz}{RGB}{219,223,238}
\definecolor{fog}{RGB}{213, 193, 234}
\definecolor{languid_lavender}{RGB}{222,203,228}
\definecolor{watusi}{RGB}{254,221,207}
\definecolor{coral_andy}{RGB}{243,204,205}
\definecolor{cosmos}{RGB}{248,209,210}
\definecolor{melon}{RGB}{254,191,181}
\definecolor{azalea}{RGB}{234, 193, 194}
\definecolor{beauty_bush}{RGB}{235, 185, 179}
\definecolor{sundown}{RGB}{249, 180, 181}
\definecolor{mona_lisa}{RGB}{246,152,134}
\definecolor{salmon}{RGB}{242,131,107}

\definecolor{summer_sky}{RGB}{58, 151, 233}
\definecolor{chateau_green}{RGB}{72, 179, 96}
\definecolor{matisse}{RGB}{25, 104, 167}
\definecolor{allports}{RGB}{31, 106, 125}
\definecolor{sun_shade}{RGB}{255, 144, 68}
\definecolor{flamingo}{RGB}{237, 88, 85}
\definecolor{studio}{RGB}{128, 91, 160}

\definecolor{maya_blue}{RGB}{102, 204, 255}
\definecolor{feijoa}{RGB}{178,223,138}
\definecolor{sushi}{RGB}{117, 168, 47}
\definecolor{norway}{RGB}{158, 194, 132}
\definecolor{japanese_laurel}{RGB}{53, 116, 40}
\definecolor{see_green}{RGB}{161,228,195}
\definecolor{monte_carlo}{RGB}{135,204,194}
\definecolor{granny_smith_apple}{RGB}{150,214,150}
\definecolor{moss_green}{RGB}{170,216,176}
\definecolor{chateau_green}{RGB}{72, 179, 96}
\definecolor{opal}{RGB}{164,207,190}
\definecolor{acapulco}{RGB}{117, 170, 148}
\definecolor{viridian}{RGB}{55, 137, 122}
\definecolor{amazon}{RGB}{56, 123, 84}
\definecolor{asparagus}{RGB}{123, 160, 91}
\definecolor{fruit_salad}{RGB}{91, 160, 94}
\definecolor{puerto_rico}{RGB}{72, 179, 150}
\definecolor{mountain_meadow}{RGB}{0, 163, 136}
\definecolor{matisse}{RGB}{25, 104, 167}
\definecolor{allports}{RGB}{31, 106, 125}
\definecolor{astral}{RGB}{55, 111, 137}
\definecolor{spring_leaves}{RGB}{46, 83, 117}
\definecolor{biscay}{RGB}{44, 62, 80}
\definecolor{midnight}{RGB}{0, 29, 50}
\definecolor{amethyst}{RGB}{153, 102, 204}
\definecolor{studio}{RGB}{128, 91, 160}
\definecolor{tapestry}{RGB}{194, 109, 132}
\definecolor{atomic_tangerine}{RGB}{255, 153, 102}
\definecolor{amber}{RGB}{255, 191, 0}
\definecolor{casablanca}{RGB}{244, 178, 84}
\definecolor{california}{RGB}{233, 140, 58}
\definecolor{tomato}{RGB}{255, 97, 56} 
\definecolor{alizarin}{RGB}{233, 58, 64}

\definecolor{linen}{RGB}{251, 239, 227}
\definecolor{double_pearl_lusta}{RGB}{253, 242, 208}
\definecolor{oasis}{RGB}{253, 242, 208}
\definecolor{milan}{RGB}{255, 254, 169}
\definecolor{texas}{RGB}{245, 232, 123}
\definecolor{maize}{RGB}{249, 212, 156}

\definecolor{turmeric}{RGB}{211, 178, 76}
\definecolor{saffron}{RGB}{249,193,62}
\definecolor{my_sin}{RGB}{255, 176, 59}
\definecolor{tree_poppy}{RGB}{246, 154, 27}
\definecolor{jaffa}{RGB}{240, 131, 58}
\definecolor{crusta}{RGB}{254, 127, 44}
\definecolor{tahiti_gold}{RGB}{223, 102, 36}
\definecolor{outrageous_orange}{RGB}{255, 100, 45}
\definecolor{safety_orange}{RGB}{254, 106, 0}

\definecolor{azalea}{RGB}{251, 196, 196}
\definecolor{oyster_pink}{RGB}{238,206,205} 
\definecolor{coral_candy}{RGB}{242,208,205} 
\definecolor{baby_pink}{RGB}{246, 194, 192}
\definecolor{petite_orchid}{RGB}{223, 157, 155}
\definecolor{apricot}{RGB}{241,140,122}
\definecolor{NY_pink}{RGB}{228,136,113}
\definecolor{carmine_pink}{RGB}{231, 76, 60}
\definecolor{deep_carmine_pink}{RGB}{236, 50, 67}

\definecolor{wewak}{RGB}{244, 143, 150}
\definecolor{light_coral}{RGB}{244, 127, 123}
\definecolor{bittersweet}{RGB}{255,111,105}
\definecolor{carnation}{RGB}{245, 80, 86}
\definecolor{flamingo}{RGB}{237, 88, 85}
\definecolor{sunset_orange}{RGB}{242,89,75}
\definecolor{ku_crimson}{RGB}{243, 0, 25}
\definecolor{amaranth}{RGB}{234,46,73}
\definecolor{valencia}{RGB}{214, 87, 70}
\definecolor{chilean_fire}{RGB}{215, 87, 44}
\definecolor{mexican_red}{RGB}{170, 41, 37}

\definecolor{napa}{RGB}{163, 154, 137}

\definecolor{athens_gray}{RGB}{236, 240, 241}
\definecolor{gallery}{RGB}{240,240,240}
\definecolor{mercury}{RGB}{230,230,230}
\definecolor{platinum}{RGB}{228,228,228}
\definecolor{silver}{RGB}{191,191,191}
\definecolor{aluminum}{RGB}{153,153,153}
\definecolor{ship_gray}{RGB}{77,77,77}
\definecolor{tuatara}{RGB}{67, 67, 67}

\definecolor{malibu}{RGB}{110, 180, 240}
\definecolor{celestial_blue}{RGB}{52, 152, 219}
\definecolor{curious_blue}{RGB}{41, 128, 185}
\definecolor{french_blue}{RGB}{0, 112, 182}
\definecolor{matisse}{RGB}{25, 104, 167}
\definecolor{shakespeare}{RGB}{85, 154, 193}
\definecolor{seagull}{RGB}{128,177,211}
\definecolor{jelly_bean}{RGB}{45, 126, 150}
\definecolor{venice_blue}{RGB}{87, 135, 105}
\definecolor{boston_blue}{RGB}{68, 147, 161}

\definecolor{turquoise}{RGB}{41,217,194}
\definecolor{java}{RGB}{2,190,196}
\definecolor{riptide}{RGB}{141,211,199}
\definecolor{mountain_meadow}{RGB}{0, 163, 136}
\definecolor{free_speech_aquamarine}{RGB}{0, 156, 114}

\definecolor{cosmic_latte}{RGB}{222, 247, 229}
\definecolor{chinook}{RGB}{163, 232, 178}
\definecolor{padua}{RGB}{121, 189, 143}
\definecolor{ocean_green}{RGB}{79, 176, 112}
\definecolor{pastel_green}{RGB}{107, 227, 135}
\definecolor{chateau_green}{RGB}{69, 191, 85}
\definecolor{RoyalBlue}{RGB}{69, 191, 85}
\definecolor{pigment_green}{RGB}{0, 175, 79}
\definecolor{fern}{RGB}{101,197,117}
\definecolor{killarney}{RGB}{56, 113, 66}

\usepackage{algorithm}
\usepackage{algorithmic}

\usepackage{pifont}
\usepackage{tikz}
\usetikzlibrary{shapes, backgrounds, positioning, fit, calc}
\usetikzlibrary{arrows.meta}
\usetikzlibrary{shapes.arrows, shapes.geometric}
\usetikzlibrary{matrix}

\renewcommand{\arraystretch}{0.86}

\makeatletter
\g@addto@macro\normalsize{%
  \abovedisplayskip 6pt plus 3pt minus 3pt%
  \belowdisplayskip \abovedisplayskip
  \abovedisplayshortskip 6pt plus3pt  minus3pt%
  \belowdisplayshortskip 6pt plus3pt minus3pt%
}

\makeatother

\title{Factual Consistency Evaluation for Text Summarization via Counterfactual Estimation}
\author{Yuexiang Xie$^1$ \quad Fei Sun$^1$ \quad Yang Deng$^{2,}$\footnotemark[1] \quad Yaliang Li$^{1,}$\footnotemark[2] \quad Bolin Ding$^1$ \\
        $^{1}$Alibaba Group\quad $^{2}$The Chinese University of Hong Kong \\ \{yuexiang.xyx, ofey.sf, yaliang.li, bolin.ding\}@alibaba-inc.com \quad ydeng@se.cuhk.edu.hk\\}

\begin{document}
\maketitle
\renewcommand*{\thefootnote}{\fnsymbol{footnote}}
\footnotetext[1]{Work done at Alibaba.}
\footnotetext[2]{Corresponding author.}
\renewcommand*{\thefootnote}{\arabic{footnote}}

\begin{abstract}

Despite significant progress has been achieved in text summarization, factual inconsistency in generated summaries still severely limits its practical applications. Among the key factors to ensure factual consistency, a reliable automatic evaluation metric is the first and the most crucial one. However, existing metrics either neglect the intrinsic cause of the factual inconsistency or rely on auxiliary tasks, leading to an unsatisfied correlation with human judgments or increasing the inconvenience of usage in practice. In light of these challenges, we propose a novel metric to evaluate the factual consistency in text summarization via counterfactual estimation, which formulates the causal relationship among the source document, the generated summary, and the language prior. We remove the effect of language prior, which can cause factual inconsistency, from the total causal effect on the generated summary, and provides a simple yet effective way to evaluate consistency without relying on other auxiliary tasks. We conduct a series of experiments on three public abstractive text summarization datasets, and demonstrate the advantages of the proposed metric in both improving the correlation with human judgments and the convenience of usage. The source code is available at \href{https://github.com/xieyxclack/factual\_coco}{https://github.com/xieyxclack/factual\_coco}.
\end{abstract}

\section{Introduction}
In recent years, significant progress has been achieved in text summarization, and with the help of deep neural networks, we can generate informative, relevant, and fluent texts~\cite{see2017get,narayan2018dont,liu2019text}.
However, it still remains a major challenge to ensure the factual consistency of the generated summary with respect to the source document~\cite{zhang2020optimizing,kryscinski2020evaluating}. 
For instance, in the annotated data released by~\citet{maynez2020faithfulness}, more than 50\% of the generated summaries are not completely consistent with the source document. 
Such factual inconsistency between source document and generated summary, also known as \textit{hallucination}, undoubtedly limits practical applications of text summarization techniques.

To ensure the factual consistency in text summarization, a reliable automatic evaluation metric is the first and the most crucial factor~\cite{goodrich2019assessing}.
The predominant automatic metrics, e.g., ROUGE~\cite{lin2004rouge} and METEOR~\cite{lavie2007meteor}, are mainly based on $n$-gram lexical-overlap and have been proven to be poorly correlated with human judgments on factual consistency~\cite{bhandari2020metrics,maynez2020faithfulness,wang2020asking}.
To better evaluate the factual consistency of summarization systems, various types of metrics have been introduced, including computing semantic similarity with pretrained model instead of $n$-gram based similarity \cite{zhang2020BERTScore,koto2020ffci}, and using auxiliary tasks such as textual entailment \cite{Falke:ACL2019:ranking,maynez2020faithfulness} and question answering \cite{Chen:AAAI2018:Semantic,Eyal:NAACL2019:question,wang2020asking,scialom2021questeval}.
However, none of them tackle this issue from the view of intrinsic cause of the factual inconsistency.
Besides, some metrics rely on auxiliary tasks (e.g., question answering), which makes these metrics costly and inconvenient.
In short, automatic evaluation for factual consistency in text summarization still remains an open research problem.

Revisiting the sequence-to-sequence (Seq2Seq) summarization models, a summary is generated according to the \textit{encoded source document} and the \textit{learned decoder}.
The information in the source document is encoded and used to ensure the factual consistency of generated summary.
The decoder, as a language model, learns the language prior from the training corpus to transform the encoded source document to an informative and fluent summary.
However, the side effects also come along with the language prior, hallucinating the inconsistency tokens due to spurious linguistic correlations learned from training corpus.
For instance, when the term \textit{green leaves} occurs frequently in the training corpus and the summarization model has learned such language prior knowledge, it could be with high probability to generate such inconsistency term \textit{green leaves}, even though the source document is about \textit{red maple leaves}. 
Similar hallucination phenomena have also been observed in other conditional text generation tasks, including image caption~\cite{hendricks2018women} and  data-to-text~\cite{filippova2020controlled}.

Shed light by the above challenges and insights, in this paper, we seek to design a simple yet effective evaluation metric, named \textbf{Co}unterfactual \textbf{Co}nsistency (denoted as \textbf{CoCo}), for text summarization. 
Different from the existing metrics, CoCo is proposed to evaluate the factual consistency of summarized texts via counterfactual estimation, and it does not rely on auxiliary tasks, which brings the convenience of usage in practice.
To be specific, with the help of causal inference~\cite{pearl2018book,yao2021survey}, we formulate the causal relationship among the source document, the generated summary, and the language prior to build up the causal graph for text summarization.
According to the built causal graph and the analysis of causal effect, we point out that the effect of language prior can be blamed to cause factual inconsistency.
Thus, we propose counterfactual abstractive summarization to estimate the causal effect of language prior on the generated summary, and remove it from the total causal effect.
The estimated effect, which is the causal effect of the source document on the generated summary, serves as a factual consistency score of the generated summary.
The intuition is that when texts are generated more relying on the source document rather than the language prior, they should be more likely to be factually consistent w.r.t. the source documents.

To demonstrate the effectiveness of the proposed metric CoCo, we conduct a series of experiments on three public datasets, which are derived from widely-used benchmarks CNN/Daily Mail~\cite{Hermann2015Advances,nallapati2016abstractive} or XSUM~\cite{narayan2018dont}, and have human annotations on factual consistency. Without relying on auxiliary tasks, the proposed metric CoCo achieves a significant improvement against the existing automatic metrics for text summarization in terms of the correlation with human annotations.

\section{Related Work}

The most popular $n$-gram based evaluation metrics, e.g., ROUGE~\cite{lin2004rouge}, BLEU~\cite{papineni2002bleu} and METEOR~\cite{lavie2007meteor}, have been proven to perform poorly on measuring factual consistency \cite{wang2020asking}. 
Inspired by the success of pretrained contextual word embeddings, BERTScore \cite{zhang2020BERTScore} leverages the pretrained BERT~\cite{devlin2019bert} model to compute the similarity between the generated summary and reference.
However, these metrics cannot lead to a satisfying correlation to human judgments on factual consistency since they only capture the token-level overlapping or similarity.
Hence, instead of defining metrics on token level, \citet{goodrich2019assessing} proposes to measure the factual consistency by counting the overlap of facts (i.e., relation tuple) extracted from the generated summary and the source document.

Several works have also explored Natural Language Inference (NLI) to evaluate the factual consistency via calculating entailment probability between the document and its abstractive summaries \cite{Falke:ACL2019:ranking,maynez2020faithfulness}.
They assume that a factually consistent summary is usually entailed by the source document.
To address the issue of domain shift in out-of-the-box NLI models, synthetic training datasets, e.g., augmented by summarization datasets \cite{kryscinski2020evaluating} and QA datasets \cite{mishra2021looking}, are created to finetune the BERT-based NLI models.

Question answering has also been used as an evaluation method for summarization \cite{Mani:EACL1999:tipster,Clarke:CL2010:discourse}.
For factual consistency evaluation, the basic intuition is to test whether a summary and its corresponding source have similar answers for the synthetic questions.
The differences between various works are mainly in question generation and metric computation.
For example, \citet{Eyal:NAACL2019:question} generates questions from the reference summary; \citet{wang2020asking} and \citet{Durmus:ACL2020:feqa} generate questions from the evaluated summary; while \citet{scialom2021questeval} generates questions from both the evaluated summary and its corresponding source document.
For metric computation, \citet{Eyal:NAACL2019:question} averages the percentage of questions answered correctly according to the generated summaries, while \citet{wang2020asking} computes the similarity of the two answers from the summary and its corresponding source with token-level F$_1$ score.

In summary, the existing metrics are proposed to measure the factual consistency via calculating the semantic similarity between the generated summary and the source document or the references, with the help of extrinsic tasks like pretrained language models or other auxiliary tasks.
However, few of them explore the intrinsic cause of the factual inconsistency. In this paper, we study this task from the perspective of causal inference~\cite{Judea2001Direct,pearl2018book}. 

\section{Methodology}
In this section, we introduce the proposed metric for measuring factual consistency in text summarization, named \textbf{Co}unterfactual \textbf{Co}nsistency (denoted as CoCo).

\subsection{Causal Graph of Text Summarization}

We first introduce two key concepts of causal inference, i.e., causal graph and causal effect.
The \textit{causal graph} represents the causal relationships between variables using a directed acyclic graph $G=\{V, E\}$, where $V$ denotes the set of variables and $E$ represents the cause-effect relationships among these variables.
Fig.~\ref{fig:causal_graph}(a) shows an example of causal graph with four variables.
In a causal graph, capital letters (e.g., $K$ and $X$) denote random variables and lowercase letters (e.g., $k, x$) denote their observed values, respectively. 
An edge means a causal-effect relationship between the parent (cause) and the child (effect), e.g., it can be denoted as $K \rightarrow Y$.

Here, we build the causal graph of abstractive text summarization as illustrated in Fig.~\ref{fig:causal_graph}(a).
It reflects the causal relationships among the fact $C$, the source document $X$, the language prior $K$, and the generated summary $Y$.
The paths $C\rightarrow X$ and $K \rightarrow X$ represent the source document $X$ (e.g., an informative and fluent news report) is composed by the fact $C$ (e.g., the happened event) and the language prior knowledge $K$.
The paths $K\rightarrow Y$ and $X \rightarrow Y$ reflect the causal relationships in the generation process of a summary $Y$,
which can be interpreted as: The encoder of a Seq2Seq model comprehends the input document $X$ and then the decoder transforms the hidden representation outputed from the encoder into a summary with the help of the language prior knowledge $K$ (e.g., the usage of demonstratives and prepositions, the logistic relationships intra- and inter-sentences, etc.).

The causal graph in Fig.~\ref{fig:causal_graph}(a) provides an insight for measuring factual consistency for text summarization.
For an abstractive text summarization model, it demands both the information of source document (i.e., the causal effect shown by $X\rightarrow Y$), as well as the language prior (i.e., the causal effect shown by $K\rightarrow Y$) to generate the summary. 
The information provided by the source document $X$ ensures the summarization model to generate an informative and relevant summary.
The language prior brings benefits such as grammar rules; however, on the other hand, it can also lead to hallucinate the inconsistency tokens via introducing spurious linguistic correlations or even biased knowledge learned from the training corpus~\cite{Hermann2015Advances,niu2020counterfactual}. 

\tikzset{
  FARROW/.style={arrows={-{Latex[length=1.25mm, width=1.mm]}}, thick},
  causal_node/.style = {circle, draw=black, minimum width=1.6em, align=center, inner sep=0, outer sep=0, font=\small, thick},
  cf_node/.style = {circle, draw=black, fill=silver!38!platinum, minimum width=1.6em, align=center, inner sep=0, outer sep=0, font=\small, thick},
  split fill/.style args={#1 and #2}{path picture={
    \fill [#1] (path picture bounding box.south west)
      rectangle (path picture bounding box.north);
    \fill [#2] (path picture bounding box.south)
      rectangle (path picture bounding box.north east);
}}
}

\begin{figure}
    \centering
    \resizebox{\linewidth}{!}{
    \begin{tikzpicture}[]
     \node [causal_node, align=center] (x) at (0,0) {$X$};   
     \node [causal_node, align=center, left of = x, node distance=1.2cm] (c)  {$C$}; 
     \node [causal_node, align=center, right of = x, node distance=1.2cm] (y) {$Y$}; 
     \node [causal_node, align=center, above of = x, node distance=1.2cm] (k) {$K$}; 
     
     \draw[FARROW] (x) -- (y);
     \draw[FARROW] (c) -- (x);
     \draw[FARROW] (k) -- (x);
     \draw[FARROW] (k) -- (y);
     
     \node [causal_node, align=center, right of = y, node distance=3.2cm] (x1) {$x$};   
     \node [causal_node, align=center, left of = x1, node distance=1.2cm] (c1)  {$c$}; 
     \node [causal_node, align=center, right of = x1, node distance=1.2cm] (y1) {$Y$}; 
     \node [causal_node, align=center, above of = x1, node distance=1.2cm] (k1) {$k$}; 
     
     \draw[FARROW] (x1) -- (y1);
     \draw[FARROW] (c1) -- (x1);
     \draw[FARROW] (k1) -- (x1);
     \draw[FARROW] (k1) -- (y1);
     
     \node [causal_node, align=center, right of = y1, node distance=2.7cm, fill=silver!38!platinum] (cx)  {$x^*$}; 
     \node [align=center, below of = cx, node distance=1cm] (ax)  {\scriptsize $X= x^*$}; 
     \node [causal_node, align=center, left of = cx, node distance=1.2cm] (cc)  {$c$}; 
     \node [causal_node, align=center, right of = cx, node distance=1.2cm] (cyp) {}; 
      \path[split fill=silver!38!platinum and white] ($(cyp)$)
         circle [radius=0.8em]; 
      \node [causal_node, align=center, right of = cx, node distance=1.2cm] (cy) {$Y$}; 
     
     \node [causal_node, align=center, above of = cx, node distance=1.2cm] (ck) {$k$}; 
     \draw[FARROW] (ck) -- (cx) node[pos=0.5] {\color{flamingo}\ding{55}};
     \draw[FARROW] (cc) -- (cx) node[pos=0.5] {\color{flamingo}\ding{55}};
     \draw[FARROW] (ck) -- (cy.north);
     \draw[FARROW] (cx) -- (cy);
     \draw[FARROW] (ax) -- (cx);
     
     \node [align=center, below of = y1, node distance=0.6cm] (ay1)  {\scriptsize $Y_{x,k}$};
     \node [align=center, below of = cy, node distance=0.6cm] (acy)  {\scriptsize $Y_{x^*,k}$};
     \node [align=center, above of = x1, node distance=0.6cm, xshift=1.95cm] (a)  {\Large \color{flamingo}\textbf{---}};
     
     \node [align=center, below of = x, node distance=1.5cm] (a)  {(a)};
     \node [align=center, below of = x1, node distance=1.5cm, xshift=1.95cm] (a)  {(b)};

    \end{tikzpicture}}
\caption{(a) The causal graph of text summarization reflects the causal relationships among the fact $C$, source document $X$, language prior $K$, and the model-generated summary $Y$. (b) According to Eq.~\eqref{eq:total_direct_effect}, the causal effect of $X$ on $Y$ can be obtained by subtracting the effect of $K$ on $Y$ from the total effect.}
\label{fig:causal_graph}

\end{figure}
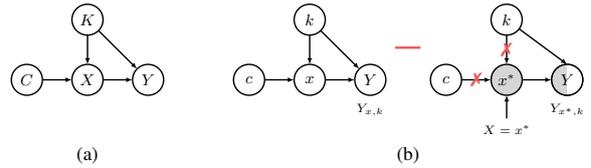

\subsection{Causal Effect}

Inspired by the intrinsic cause of factual consistency in text summarization discussed above, we propose a novel automatic evaluation metric, named \textbf{Co}unterfactual \textbf{Co}nsistency (denoted as CoCo), to measure the factual consistency via counterfactual estimation. 
To be specific, we aim to estimate the causal effect of $X\rightarrow Y$ to measure the factual consistency of the generated summary, since when the summaries are generated relied on the source document, they should be more likely to be factually consistent w.r.t. the source document than those generated relied on language prior.

To achieve this, we first need to use counterfactual notations to translate causal assumptions from graphs to formulas.
For causal graph in Fig.~\ref{fig:causal_graph}(a), given that if $C$ is set to $c$ and $K$ is set to $k$, the value that summary $Y$ could be is denoted as:
\begin{equation}
Y_{c,k} = Y(\text{do}(C=c), \text{do}(K=k)).
    \label{eq:co_no}
\end{equation}
Since there is no confounder of $C$ and $K$, we have that $\text{do}(C=c)$ equals to $C=c$ and $\text{do}(K=k)$ equals to $K=k$.
Without loss of generality, we omit the do operator for simplicity.
Thus, Eq.~\eqref{eq:co_no} can be simplified as:
\begin{equation}
Y_{c,k} = Y(C=c, K=k).
    \label{eq:co_simp}
\end{equation}
Similarly, the counterfactual notation of source document $X$ can be $X_{c,k}=X(C=c, K=k)$.

As shown in Fig.~\ref{fig:causal_graph}(a), there exist two paths directly connected to $Y$, i.e., $K{\rightarrow} Y$ and $X{\rightarrow} Y$.
Thus, $Y_{c,k}$ can be rewritten as the function of $K$ and $X$:
\begin{equation}
    Y_{c,k} = Y_{x,k} = Y(X=x, K=k).
\end{equation}
In the factual scenario, we have $x = X_{c,k} = X(C=c, K=k)$.
While, in the counterfactual scenario, $k$ can be set as different values.
For example, $Y_{x,k^*}$ describes the situation $K$ is set to $k^*$. The term ``$*$'' denotes the no-treatment condition, which can be interpreted as eliminating the causal effect via setting the variable to an empty value.
Note that such case can only happen in the counterfactual scenario.

To measure \textit{causal effects}, we need to compare two potential outcomes of the same individual given two different treatments.
For example, comparing the outcomes of taking the drug (i.e., treatment) and not taking the drug (i.e., no-treatment) to estimate the effect of a drug on a disease. 
Here, to estimate the causal effect of input source document $X$ on summary $Y$, we aim to compare the summaries of feeding document $x$ to the summarization model (i.e., $X{=}x{=}X(c,k)$), and the document $x$ is not given (i.e., the no-treatment condition $X=x^*=X(c^*,k^*)$.
As shown in Fig.~\ref{fig:causal_graph}(a), the outcome of $Y$ is also affected by the language prior knowledge $K$. Thus we should take both $X$ and $K$ into consideration when estimating the effect of $X$ on $Y$.
However, estimating the effect of $X$ on $Y$ via $Y_{x,k^*} - Y_{x^*,k^*}$ is impractical, since it is hard to block the effect of $K$ (i.e., $K=k^*$) for an abstractive summarization system.

To address this issue, we propose to estimate the causal effect of $X$ on $Y$ by subtracting the causal effect of $K$ on $Y$ from the total effect.
The total effect of treatment $X = x$ and $K=k$ on $Y=y$ compares hypothetical situations ($X = x$, $K = k$) and ($X = x^*$, $K = k^*$), which is denoted as
\begin{equation}
    E_\text{total} = Y_{x,k} - Y_{x^*,k^*},
\end{equation}
which represents the difference between the output of taking the treatment (i.e., $Y_{x,k}$) and that of no-treatment condition (i.e., $Y_{x^*,k^*}$).

For the causal effect of $K$ on $Y$, we propose counterfactual abstractive summarization to estimate it by blocking the effect of $X$.
Counterfactual abstractive summarization describes the scenario where $K=k$ and $X$ had been $x^*$. 
Since the response of $X$ is blocked, the model can only rely on the given language prior $k$ to generate summaries.
Thus, the causal effect of $K$ on $Y$ can be obtained by comparing counterfactual abstractive summarization to the no-treatment conditions:
\begin{equation}
    E_{\text{K}} = Y_{x^*,k} - Y_{x^*,k^*},
\end{equation}
where $E_{\text{K}}$ denotes the changes in the outcome $Y$ with $K$ changing from $k^*$ to $k$ and $X$ is set to the value it would have obtained at $X=x^*$.

Thus, the causal effect of $X$ on $Y$, denoted as $E_X$, can be obtained by subtracting $E_{\text{K}}$ from $E_\text{total}$:
\begin{equation}
        E_X =  E_\text{total} -  E_{\text{K}} = Y_{x,k} - Y_{x^*,k}.
    \label{eq:total_direct_effect}
\end{equation}
It can be observed that Eq.~\eqref{eq:total_direct_effect} is equivalent to the comparison between the generated summaries of conventional abstractive summarization and counterfactual abstractive summarization, as illustrated in Fig.~\ref{fig:causal_graph}(b), which implies the approach to measure the factual consistency in text summarization. 
The term $Y_{x,k}$ happens to be a standard abstractive summarization model that takes $x$ as input and outputs $y$, with the help of language prior $k$ learned from training corpus; while the term $Y_{x^*,k}$ describes a model that generates $y$ only depend on the language prior $k$, in normal case it works like a language model.

\begin{table*}[t]
	\centering
	\begin{adjustbox}{max width=0.98\textwidth}
	\begin{tabular}{p{1.8cm}p{10cm}p{5cm}}
		\toprule
		Operator & Source document $X$  & Summary $Y$ \\
		\midrule
	 ~~ \newline Sentence-level mask	& \hl{People with a DNA variation in a gene called PDSS2 tend to drink fewer cups of \mbox{\uwave{coffee}}, a study carried out at the University of Edinburgh has found.} It suggests the gene reduces cell ability to break down caffeine$\ldots$ 
		 &
		 Researchers have identified a gene that appears to curb  \uwave{coffee} consumption.\\
		 \midrule
	~~ \newline Span-level mask	&  People with a DNA variation \hl{in a \mbox{\uwave{gene}} called PDSS2} tend to drink fewer cups of coffee, a study carried out at the University of Edinburgh has found. It \hl{suggests the \mbox{\uwave{gene}} reduces cell} ability to break down caffeine$\ldots$ 
		 &
		 Researchers have identified a \uwave{gene} that appears to curb  coffee consumption.\\
		\bottomrule
	\end{tabular}
	\end{adjustbox}
	\caption{Two examples to show the mask operation on the source document $X$ according to summary $Y$. We mark the contents (\hl{green background}) that are relevant to \uwave{\textit{coffee}} and \uwave{\textit{gene}} with different strategies for demonstration.}
	\label{table:example}
\end{table*}

\subsection{CoCo Metric}
There exist several ways to implement Eq.~\eqref{eq:total_direct_effect} by applying different functions to approximate $Y_{x,k} - Y_{x^*,k}$, such as lexical overlapping and semantic similarity between the generated summaries. 
For text summarization, given source document $X$, the outputs of the model are the probability distributions $\Pr(\cdot|X)$ over the vocabulary, from which a series of tokens are sampled to compose a summary.
Thus, from another point of view, functions that can be applied to the probability distribution are also suitable to approximate $Y_{x,k} - Y_{x^*,k}$, such as the perplexity and uncertainty~\cite{xu2020understanding,xiao2021hallucination}. 
In this study, taking both the effectiveness and convenience into consideration, we adopt the probabilities of the tokens of evaluated summaries, i.e., $\Pr(y_t)\ \forall y_t \in Y$, to implement Eq.~\eqref{eq:total_direct_effect} as our automatic evaluation metric CoCo for factual consistency.
Besides, we adopt an independent summarization model as the \textit{scoring model} in the instantiation of CoCo, rather than using the model that generates the evaluated summary, considering that the factual consistency can be biased by the model that produced this evaluated summary.
By adopting an independent summarization model as the scoring model, CoCo can be applied to score any given summary based on its corresponding source document, disregarding how the evaluated summary is generated.

In practice, key tokens, e.g., nouns and verbs, usually play a more important role in measuring the factual consistency than conjunctions and symbols. 
However, the final score can be easily dominated by the probability of those negligible tokens (e.g., conjunctions) and becomes meaningless since they often have a much higher generation probability than the key tokens.
To address this issue, we only count the probability of key tokens (denoted as $Y'$) in the evaluated summary. 
The criteria of selecting key tokens can be task-oriented designed, e.g., one can select disease names as key tokens for measuring the factual consistency of radiology reports~\cite{zhang2020optimizing}.
It is worth noting that here we do not remove other tokens in the evaluated summary, we just ignore their scores.

Assume that a scoring model takes the source document $x=\{w_1,w_2,\dots, w_n\}$ as input, and tries to generate the evaluated summary in an auto-regressive manner.
At $t$-th step during decoding, the scoring model outputs a probability distribution among the vocabulary $\Pr(\cdot | X, y_{<t})$.
For the term $Y_{x^*,k}$ in Eq.~\eqref{eq:total_direct_effect}, a natural choice is to obtain the probability of $Y'$ with the empty source document $x^*$.
However, in practical use, empty input could make the summarization model into an ill-posed state that produces almost zero probability for all tokens in the evaluated summary due to the mismatch between training and inference.
To tackle this issue, we perform mask operation on the relevant content in $x$ that is considered as what the model relies on during the generation process of $Y'$. 
For each token in $Y'$, the masked content could be token-level, span-level, sentence-level, and even the whole document (different mask strategies are evaluated in the experiment section). We show two examples of how we mask the source document $X$ according to $Y$ in Table~\ref{table:example}. 
\begin{algorithm}[t]
	\caption{CoCo metric}
	\label{algorithm1}
	\begin{algorithmic}[1]
		\REQUIRE Source document $X$, model-generated summary $Y$, scoring model $F$
		\ENSURE Factual consistency score of $Y$ w.r.t. $X$ (i.e., the CoCo value)
		\STATE Select key tokens $Y'$ from $Y$;
		\STATE Mask the source document $X$ according $Y'$ to produce $X'$;
		\STATE Feed $X$ and $X'$ into the scoring model $F$ respectively to generate the probability of each tokens in $Y'$, i.e., $\Pr(y_i|X,y_{<i})$ and $\Pr(y_i|X',y_{<i})$, $\forall y_i\in Y'$ ;
		\STATE Calculate the CoCo value according to Eq.~\eqref{eq:coco}.
	\end{algorithmic}
\end{algorithm}

Finally, we aggregate the mask content according to all the tokens in $Y'$ and result in a mask source document $X'$.
The masked source document $X'$ is fed into the scoring model and the decoder produces another probability distribution $\Pr(\cdot | X', y_{<t})$ accordingly. The definition of CoCo can be formally given as:
\begin{equation}
\small
	\text{CoCo} = \frac{1}{|Y'|}\sum_{y_t\in Y'}\Pr(y_t | X, y_{<t}) - \Pr(y_t | X', y_{<t}),
	\label{eq:coco}
\end{equation}
where $y_{<t}$ denotes the all prefix tokens in $Y$ before $t$-th step\footnote{when $y_t$ is the first token in $Y$, $y_{<t}=\emptyset$.}, and $\Pr(y_t | X, y_{<t})$ and $\Pr(y_t | X', y_{<t})$ represents the predicted probability of token $y_t$ at $t$-th when given the prefix tokens and the source document $X$ or its masked version $X'$.
The algorithm of CoCo is illustrated in Algorithm~\ref{algorithm1}.

\section{Experiments}
\subsection{Datasets}
To evaluate the effectiveness of the proposed metric, we conduct experiments on three public abstractive text summarization datasets.

\noindent \textbf{QAGS-CNN/DM.} It is a subset of CNN/Daily Mail dataset~\cite{Hermann2015Advances,nallapati2016abstractive} and released by~\citet{wang2020asking}. This dataset contains 235 instances collected from the test set of CNN/Daily Mail, and each instance consists of a source document, a reference summary, and a model-generated summary produced by a bottom-up abstractive summarization model~\cite{gehrmann2018bottom}. The generated summaries are assigned with human annotations to indicate their factual consistency scores.

\noindent \textbf{QAGS-XSUM.} This dataset is derived from XSUM~\cite{narayan2018dont} and contains 239 source documents, the corresponding reference summaries, and the synthetic summaries generated by BART~\cite{lewis2020bart}. \citet{wang2020asking} collects human evaluation scores on factual consistency for each generated summary via ParlAI~\cite{miller2017parlai} .

\noindent \textbf{SUMMEVAL.} It is released by~\citet{Fabbri:TACL21:SummEval} and contains the generated summaries from 16 abstractive and extractive models of 100 test data from DNN/Daily Mail. We adopt 1200 abstractive text summaries and extract 3600 public expert annotations on factual consistency for them.

\subsection{Baselines}
We adopt widely-used automatic evaluation metrics for comparison, such as ROUGE~\cite{lin2004rouge}, BLEU~\cite{papineni2002bleu} and METEOR~\cite{lavie2007meteor}. 
For ROUGE, we adopt ROUGE-1, ROUGE-2 and ROUGE-L, which denotes that the overlapping units in calculation are set to be uni-grams, bi-grams and longest-common subsequence respectively.

Besides, BERTScore~\cite{zhang2020BERTScore}, FFCI~\cite{koto2020ffci}, QAGS~\cite{wang2020asking}, and QuestEval~\cite{scialom2021questeval} are also adopted as baselines in the experiment.
We use the outputs from the $17$-th layer in \texttt{roberta-large} to implement BERTScore as suggested by~\citet{zhang2020BERTScore}. 
For FFCI, a framework that can be implemented by different basic metrics, we use FFCI$_\text{ROUGE-1}$, FFCI$_\text{ROUGE-2}$, FFCI$_\text{ROUGE-L}$, FFCI$_\text{BERTScore}$ to distinguish the different basic metrics as the original paper suggested. 
For QA-based metrics QAGS and QuestEval, following the settings in the original papers, the QG model and QA model in QAGS are implemented based on BART and BERT respectively, and for QuestEval, they are both implemented based on T5~\cite{Colin2020T5}. 
And we adopt QuestEval$_\text{precision}$, QuestEval$_\text{recall}$, QuestEval$_{\text{F}_1}$ for calculating the precision, recall, and F$_1$ score of the answers given the source documents and generated summaries. 

For the proposed metric CoCo, we investigate different mask strategies in our study. We use CoCo$_\text{token}$, CoCo$_\text{span}$, CoCo$_\text{sent}$, and CoCo$_\text{doc}$ to denote token-level, span-level (i.e., to mask five successive tokens that contain the key token as the center), sentence-level and document-level (i.e., to mask the whole document) mask strategies when calculating the causal effect of the source document on the generated summary.

\subsection{Implementation}
We adopt BART~\cite{lewis2020bart} as the scoring model for the proposed metric CoCo. To be specific, we feed the whole source document or masked source document into the encoder of BART and apply the teacher forcing~\cite{Bengio2015Scheduled} during the decoding process. At $t$-th step of decoding, we take the output probability of the $t$-th token in the evaluated summary $Y$, i.e., $\Pr(y_t)$, as one of the factual consistency scores for the evaluated summary if $y_t$ is a key token recognized by part-of-speech tagging toolkit \texttt{spaCy}\footnote{https://spacy.io/}, otherwise, we discard it. 

In our study, the BART model is finetuned on the training set of CNN/Daily Mail datasets for the experiments conducted on QAGS-CNN/DM and SUMMEVAL, and finetuned on the training set of XSUM for QAGS-XSUM.  For baseline models, the implementation is based on the huggingface~\cite{wolf2020transformers}. More details of the implementation can be found in the Appendix. All models are implemented using \texttt{PyTorch}~\cite{Paszke2019PyTorch} and performed on GeForce RTX 2080 Ti GPUs.

\subsection{Comparison Results}
\begin{table*}[t]
\renewcommand{\arraystretch}{0.85}
	\centering
	\setlength{\tabcolsep}{1em}
	\begin{tabular}{l rrrr rr}
		\toprule
		\multirow{2}{*}{Metric} & \multicolumn{2}{c}{QAGS-CNN/DM} & \multicolumn{2}{c}{QAGS-XSUM} & \multicolumn{2}{c}{SUMMEVAL} \\ \cmidrule(lr){2-3} \cmidrule(lr){4-5} \cmidrule(lr){6-7}
		&  \multicolumn{1}{c}{$r$} & \multicolumn{1}{c}{$\rho$} & \multicolumn{1}{c}{$r$} & \multicolumn{1}{c}{$\rho$} & \multicolumn{1}{c}{$r$} & \multicolumn{1}{c}{$\rho$} \\
		\midrule
		ROUGE-1 & 29.01 & 23.87 & 13.12 & 12.82 & 20.23 & 17.72 \\
		ROUGE-2 & 17.91 & 18.78 & 8.66 & 9.96 & 16.72 & 16.72 \\
		ROUGE-L & 23.43 & 24.09 & 8.36 & 10.66 & 19.20 & 18.16\\
		METEOR & 25.65 & 24.56 & 10.78 & 11.59 & 16.91 & 14.38\\
		BLEU & 17.63 & 22.85 & 2.55 & 2.55 & 10.83 & 10.73\\
		\midrule
		BERTScore & 37.41 & 36.36 & 11.25 & 13.23 & 18.58 & 17.53\\
		FFCI$_\text{ROUGE-1}$ & 43.55 & 42.00 & 13.70 & \underline{19.26} & 36.46 & 33.86\\
		FFCI$_\text{ROUGE-2}$ & 45.01 & 43.62 & 18.96 & 18.63 & 37.95 & \underline{35.40}\\
		FFCI$_\text{ROUGE-L}$ & 43.11 & 41.32 & 16.67 & 17.54 & \underline{38.02} & 34.45\\
		FFCI$_\text{BERTScore}$ & 48.47 & \underline{48.62} & \underline{20.04} & 19.04 & 28.54 & 30.76 \\
		QAGS & 31.39 & 35.96 & 15.18 & 17.48 & 17.71 & 12.65\\
		QuesEval$_\text{precision}$ & 38.02 & 35.96 & 5.66 & 7.43 & 33.53 & 29.52\\
		QuesEval$_\text{recall}$ & 41.10 & 36.40 & 6.57 & 7.33 & 26.95 & 25.69\\
		QuesEval$_{\text{F}_1}$ & \underline{49.18} & 44.53 & 7.03 & 9.63 & 36.96 & 33.93\\
		\midrule
		CoCo$_\text{token}$ (ours) & 55.79 & 49.33 & 19.02 & 19.83 & 42.67 & 40.09\\
		CoCo$_\text{span}$ (ours) & 57.28 & 50.14 & 18.71 & 18.65 &   \textbf{43.57} &  \textbf{40.96}\\
		CoCo$_\text{sent}$ (ours) & \textbf{58.84} &  \textbf{52.25} &  \textbf{24.08} &  \textbf{22.70} & 42.04 & 39.03\\
		CoCo$_\text{doc}$ (ours) & 55.27  & 49.54 & 19.57 & 18.21 & 41.18 & 39.61 \\
		\bottomrule
	\end{tabular}
	\caption{Pearson correlation (denoted as $r$) and Spearman correlation (denoted as $\rho$) between automatic metrics and human judgments of factual consistency on text summarization datasets. The bold scores are the best among all the metrics, while the underlined scores are the best among the baseline metrics. }
	\label{table:exp_result}
\end{table*}

We report the Pearson correlation coefficient ($r$) and Spearman correlation coefficient ($\rho$) between various automatic evaluation metrics and human judgments on factual consistency in Table~\ref{table:exp_result} (results are shown as percentages). 
The larger value represents the better positive correlation with human judgments, which demonstrates the effectiveness of the automatic evaluation metric on factual consistency for text summarization.

From the table, we can observe that the $n$-gram based metrics, including ROUGE, METEOR, and BLEU, achieve the worse results on three datasets compared with most of other the baselines. These results are consistent with the observation in previous studies \cite{bhandari2020metrics,maynez2020faithfulness}. 
Although convenient for adopting, $n$-gram based metrics are poorly correlated with human judgments on factual consistency, since they only care about the lexical overlap between the generated summary and reference, but fail to capture the semantic similarity.

Compared to $n$-gram based metrics, BERTScore could not achieve consistent improvements on three datasets (better on QAGS-CNN/DM but slightly worse on QAGS-XSUM and SUMMEVAL). And the comparisons between FFCI and its corresponding basic metric (e.g., FFCI$_\text{BERTScore}$ v.s. BERTScore) show the advantages brought by calculating the lexical-overlap or semantic similarity between the generated summaries and the source document rather than the reference when measuring factual consistency. However, since FFCI adopts a sentence-level evaluation manner, the cost can be nearly $m \times n$ times of other metrics when a document and a generated summary containing $m$ and $n$ sentences respectively.

On the other hand, QA-based metrics QAGS and QuestEval introduce auxiliary tasks, including question generation (QG) and question answering (QA), to evaluate factual consistency. Both of them achieve competitive results compared with other baselines. However, QA-based metrics highly rely on well-trained QA and QG models, which makes them inconvenient, or unavailable in languages without enough corpus for pretraining. Meanwhile, it is worth pointing out that these metrics are computationally expensive~\cite{koto2020ffci}, and easily lead to error propagation and accumulation because of their pipeline manner.

The proposed metric CoCo outperforms other baselines by a noticeable margin on three adopted datasets. For instance, the Pearson correlation coefficient between CoCo$_\text{sent}$ and human judgments on QAGS-CNN/DM is 58.84, which is significantly larger than those of the best result among baselines (49.18 for QuesEval$_{\text{F}_1}$), and is twice times of the best result among $n$-gram based metrics (29.01 for ROUGE-1).
These results demonstrate the effectiveness of CoCo in improving the correlation with human judgments. 
In the bottom subgroup of Table~\ref{table:exp_result}, we report the comparison results among different mask strategies applied in CoCo. 
From these results, we can see the span-level and sentence-level mask strategies are better than token-level and document-level, which implies that a suitable mask range for relevant content is important for measuring factual consistency. 
The too-small mask range could cause that the decoder is still able to infer the masked tokens from the context, while the too-large mask range might weaken the effect of language prior and lead to near-zero scores for all the tokens in the evaluated summary.

\subsection{Further Discussions}

In this section, we provide some discussions about the proposed CoCo metric to better understand how it works to measure the factual consistency of the model-generated summary.

\begin{figure}
	\centering
	\includegraphics[width=0.48\textwidth]{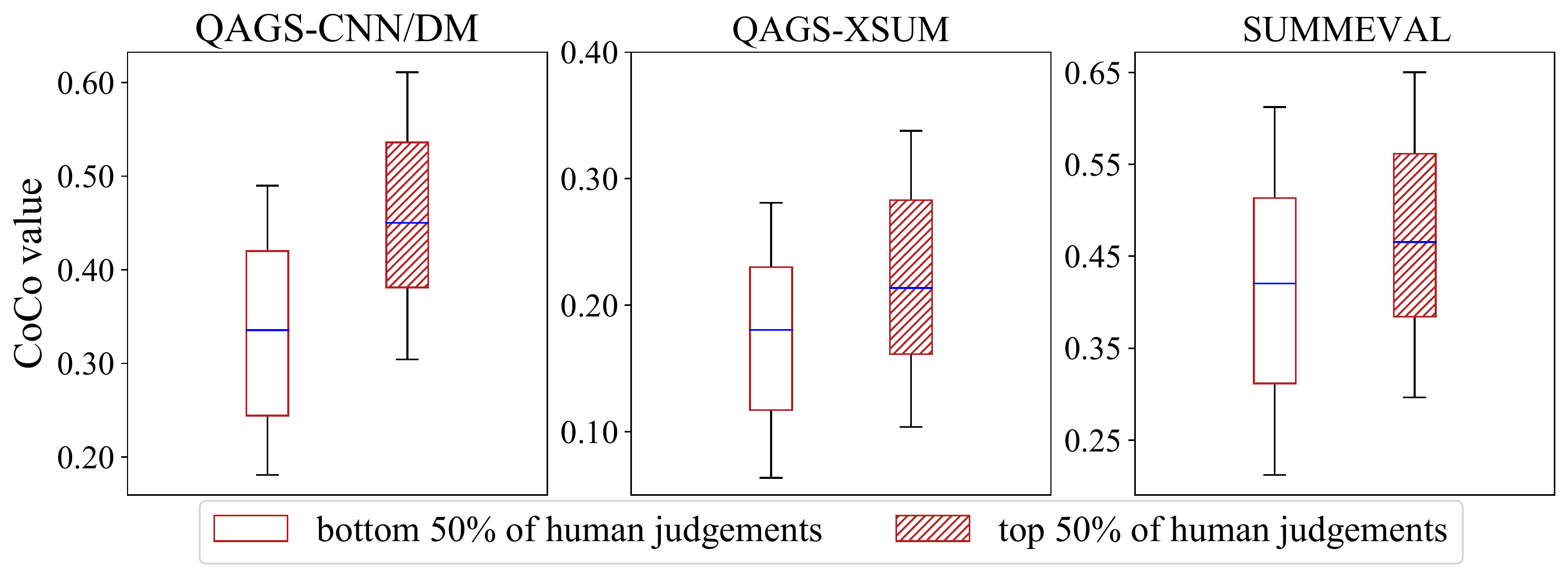}
	\caption{Comparison of CoCo values assigned to high (top 50\%) and low (bottom 50\%) human judgments.}
	\label{figure:human_def}
\end{figure}

\noindent \textbf{CoCo values w.r.t. various human judgments.} For each adopted dataset, we split the instances into two halves according to the human judgments assigned to the model-generated summaries. Then we calculate the statistics of CoCo values for each half respectively and illustrate the results in Fig.~\ref{figure:human_def}. 
It can be observed that summaries with high human judgments (i.e., in the top 50\%) are also assigned with high CoCo values, which demonstrates CoCo values are consistent with human judgments on factual consistency. Thus the proposed metric CoCo is a reliable automatic evaluation metric for factual consistency in text summarization.

\noindent \textbf{Different scoring models.}
We adopt four different scoring models to implement CoCo, including BART, BERT~\cite{liu2019text}, T5~\cite{Colin2020T5}, and PEGASUS~\cite{zhang2020pegasus} (denoted as PEGA). 
The experiment results are shown in Table~\ref{tab:exp_diff_scoring_model}, which demonstrate that among the four adopted scoring models, CoCo achieves consistent and competitive results against the performance of baselines reported in Table~\ref{table:exp_result}.

\begin{table}[t]
    \centering
    \renewcommand{\arraystretch}{1.05}
    \setlength{\tabcolsep}{0.3em}
	\begin{tabular}{lcccc}
        \toprule
        Datasets & BART & BERT & T5 & PEGA \\
        \midrule
         QAGS-CNN/DM & 58.84 & 53.27 & 56.13 & 58.96 \\
         QAGS-XSUM & 24.08 & 17.88 & 21.66 & 19.83 \\
         SUMMEVAL & 42.04 & 39.36 & 42.98 & 43.38 \\
         \bottomrule
    \end{tabular}
    \caption{The experiment results of Pearson correlation $r$ between CoCo and human judgments. CoCo is implemented with different scoring models shown in the header of the horizontal axis, and sentence-level mask strategy is used.}
    \label{tab:exp_diff_scoring_model}
\end{table}

\noindent \textbf{Case study.} To better understand how CoCo works, a case study is illustrated in Table~\ref{table:case_study}.
We take three tokens \textit{American}, \textit{Augusta}, and \textit{Monday} as examples and show the CoCo values. 
The tokens \textit{American} is factually inconsistent w.r.t. the source document since there does not exist any context to infer \textit{Tiger Woods} is an \textit{American} in the source document.  \textit{Augusta} and \textit{Monday} are factually consistent as they can be directly explained by the source document.

We can observe that both \textit{Augusta} and \textit{Monday} are assigned with a high CoCo value (0.4115 and 0.6346 respectively), while that of \textit{American} is significant lower (0.0251).
Such differences are caused by that both \textit{Augusta} and \textit{Monday} are generated more relied on the source document, but \textit{American} is more likely to be hallucinated by the decoder, with the help of the language prior knowledge from the training corpus.
The proposed metric CoCo can assign these hallucinations with low scores via counterfactual estimation to measure the factual consistency of the generated summary.

\begin{table}
	\centering
	\renewcommand{\arraystretch}{1.05}
	\begin{tabular}{p{7.4cm}}
		\toprule
		\textbf{Source document}\\
		\textcolor{tea_green!300}{Tiger Woods} declared himself ready to compete for a fifth Masters title after completing 11 holes of practice at \textcolor{summer_sky}{Augusta} National on \textcolor{summer_sky}{Monday}.$\ldots$\\
		\midrule
		\textbf{Summary with CoCo values}\\
		The \textcolor{flamingo}{American(0.0251)} completed 11 holes of practice at \textcolor{summer_sky}{Augusta(0.4115)} on \textcolor{summer_sky}{Monday(0.6346)}.\\
		\bottomrule
	\end{tabular}
	\caption{Case Study. The factually inconsistent token \textcolor{flamingo}{\textit{American}} is assigned with a low CoCo value since it is more likely to hallucinated from 	\textcolor{tea_green!300}{\textit{Tiger Woods}}, while both \textcolor{summer_sky}{\textit{Augusta}} and \textcolor{summer_sky}{\textit{Monday}} are assigned with high CoCo values as they are generated more relied on the source document rather than the language prior. Best viewed with color.}
	\label{table:case_study}
\end{table}
\section{Conclusions}

In this paper, we introduce CoCo, an effective and convenient metric for measuring the factual consistency of abstractive summarization models.
Inspired by the intrinsic cause of factual inconsistency in text summarization, CoCo can evaluate factual consistency via counterfactual estimation without relying on other extrinsic tasks.
Experiments on human-annotated datasets verify the advantages of our proposed metric in terms of the correlation with human judgments.
In the future, several directions can be explored. 
First, although this paper focuses on evaluation metric, the proposed idea can be incorporated into abstractive summarization models to enhance factual consistency.
Another interesting direction is to apply this metric in other conditional text generation tasks such as image caption and data-to-text generation.

\normalem
\bibliography{factual}
\bibliographystyle{acl_natbib}
\newpage
\appendix

\section{Implementation Details}
We introduce the implementation details of baselines for reproduction, including:

\noindent \textbf{N-gram based metrics}. We adopt the widely-used open source packages to implement the n-gram based metrics, including ROUGE\footnote{https://github.com/andersjo/pyrouge}, BLEU\footnote{https://pypi.org/project/bleu/} and METEOR\footnote{https://github.com/salaniz/pycocoevalcap}.
And we use the CoreNLP\footnote{https://github.com/stanfordnlp/CoreNLP} as the tokenizer.

\noindent \textbf{BERTScore}. It is implemented based on the source code released by~\citet{zhang2020BERTScore}\footnote{https://github.com/Tiiiger/bert\_score}. Following the original paper, we use the outputs from the $17$-th layer in  \texttt{roberta-large}. We also try to use \texttt{bert-large-nli} suggested by~\citet{koto2020ffci}, but it fails to perform better in the experiments.

\noindent \textbf{FFCI}. FFCI~\cite{koto2020ffci} is a framework that can be implemented by different basic metrics. The adopted basic metrics in our study including ROUGE-1, ROUGE-2, ROUGE-L, and BERTScore, whose implementation details have been introduced above. The hyperparameters are set as the paper suggested.

\noindent \textbf{QAGS}. Following the settings in~\citet{wang2020asking}, the question generation (QG) model and question answering (QA) model in QAGS are implemented based on BART and BERT respectively using the source code provided by fairseq~\cite{ott2019fairseq}. For each summary, we extract 10 named entities via \texttt{spaCy}\footnote{https://spacy.io/}, and generate totally 100 questions based on these named entities using QG model, from which 20 questions with high generated probabilities are selected. These questions are fed into a QA model, together with the corresponding source documents or the summaries, to generate answers for comparisons.

\noindent \textbf{QuestEval}. For QuestEval~\cite{Scialom:EMNLP2019:Answers}, both QG model and QA model are implemented based on T5~\cite{Colin2020T5} with the help of huggingface~\cite{wolf2020transformers}, following the original paper. We adopt the suggested settings for hyperparameters of QuestEval.

\noindent \textbf{CoCo}. We adopt four different scoring models to implement CoCo, including BART~\cite{lewis2020bart}, BERT~\cite{liu2019text}, T5~\cite{Colin2020T5}, and PEGASUS~\cite{zhang2020pegasus}. Our implementation is based on fairseq and huggingface. 

\section{Dataset Availability}

\noindent \textbf{QAGS-CNN/DM \& QAGS-XSUM}. They are released by~\citet{wang2020asking}\footnote{https://github.com/W4ngatang/qags/tree/master/data}, and annotated on Amazon Mechanical Turk3\footnote{https://www.mturk.com/} via ParlAI~\cite{miller2017parlai}.
Please refer to the original paper for more details about the annotation protocol.

\noindent \textbf{SUMMEVAL}. It is released by~\citet{Fabbri:TACL21:SummEval}\footnote{https://github.com/Yale-LILY/SummEval}. 
The expert annotations are adopted in our study as the paper suggested.
For each source document in this dataset, there exists one original reference from CNN/DailyMail dataset and 10 additional crowdsources reference summaries. 
We only use the original reference in our study.

\end{document}